\documentclass[10pt,twocolumn,letterpaper]{article}

\usepackage[accsupp]{axessibility}  

\usepackage{bm}

\usepackage{iccv}
\usepackage{times}
\usepackage{epsfig}
\usepackage{graphicx}
\usepackage{amsmath}
\usepackage{amssymb}

\usepackage[pagebackref=true,breaklinks=true,letterpaper=true,colorlinks,bookmarks=false]{hyperref}

\iccvfinalcopy 

\def\iccvPaperID{11650} 

\ificcvfinal\fi

\usepackage{times}
\usepackage{epsfig}
\usepackage{graphicx}
\usepackage{float}
\usepackage{wrapfig}
\usepackage{amsmath,amssymb,amsthm}
\usepackage{algorithm,algorithmicx,algpseudocode}
\usepackage{bm,xspace}
\usepackage{comment}
\usepackage{verbatim}
\usepackage{multirow}
\usepackage{booktabs}
\usepackage{balance}
\usepackage{url}
\usepackage{etoolbox,siunitx}
\usepackage{calc}
\usepackage{pifont,hologo}
\usepackage{nicefrac}
\usepackage{lipsum}

\usepackage[table]{xcolor}

\newcommand{\figref}[1]{Fig.~\ref{#1}}

\newcommand{\ra}[1]{\renewcommand{\arraystretch}{#1}}

\usepackage[super]{nth}
\usepackage{nicefrac}
\robustify\bfseries

\begin{document}

\title{BiFF: Bi-level Future Fusion with Polyline-based Coordinate \\ for Interactive Trajectory Prediction}

\author{Yiyao Zhu \quad Di Luan \quad Shaojie Shen \\
The Hong Kong University of Science and Technology \\
{\tt\small \{yzhucp, dluan\}@connect.ust.hk, eeshaojie@ust.hk}
}

\maketitle
\ificcvfinal\thispagestyle{empty}\fi

\begin{abstract}
   Predicting future trajectories of surrounding agents is essential for safety-critical autonomous driving. Most existing work focuses on predicting marginal trajectories for each agent independently. However, it has rarely been explored in predicting joint trajectories for interactive agents. In this work, we propose Bi-level Future Fusion (BiFF) to explicitly capture future interactions between interactive agents. Concretely, BiFF fuses the high-level future intentions followed by low-level future behaviors. Then the polyline-based coordinate is specifically designed for multi-agent prediction to ensure data efficiency, frame robustness, and prediction accuracy. Experiments show that BiFF achieves state-of-the-art performance on the interactive prediction benchmark of Waymo Open Motion Dataset.
   
\end{abstract}

\section{Introduction}
\label{sec:Intro}

Motion prediction is crucial for intelligent driving systems as it plays a vital role in enabling autonomous vehicles to comprehend driving scenes and make safe plans in interactive environments. Predicting the future behaviors of dynamic agents is challenging due to the inherent multi-modal behavior of traffic participants and complex scene environments. Learning-based approaches \cite{alahi2016slstm, casas2020implicit, chai2019multipath, cui2019multimodal, deo2018cslstm, gao2020vectornet, zhao2020tnt, ding2019predicting} have recently made notable strides in this area.  By utilizing large-scale real-world driving datasets \cite{caesar2020nuscenes, ettinger2021large}, learning-based frameworks can effectively model complex interactions among agents and considerably enhance the accuracy of motion prediction.

\begin{figure}[t]
    \centering
    \includegraphics[width=1.0\columnwidth]{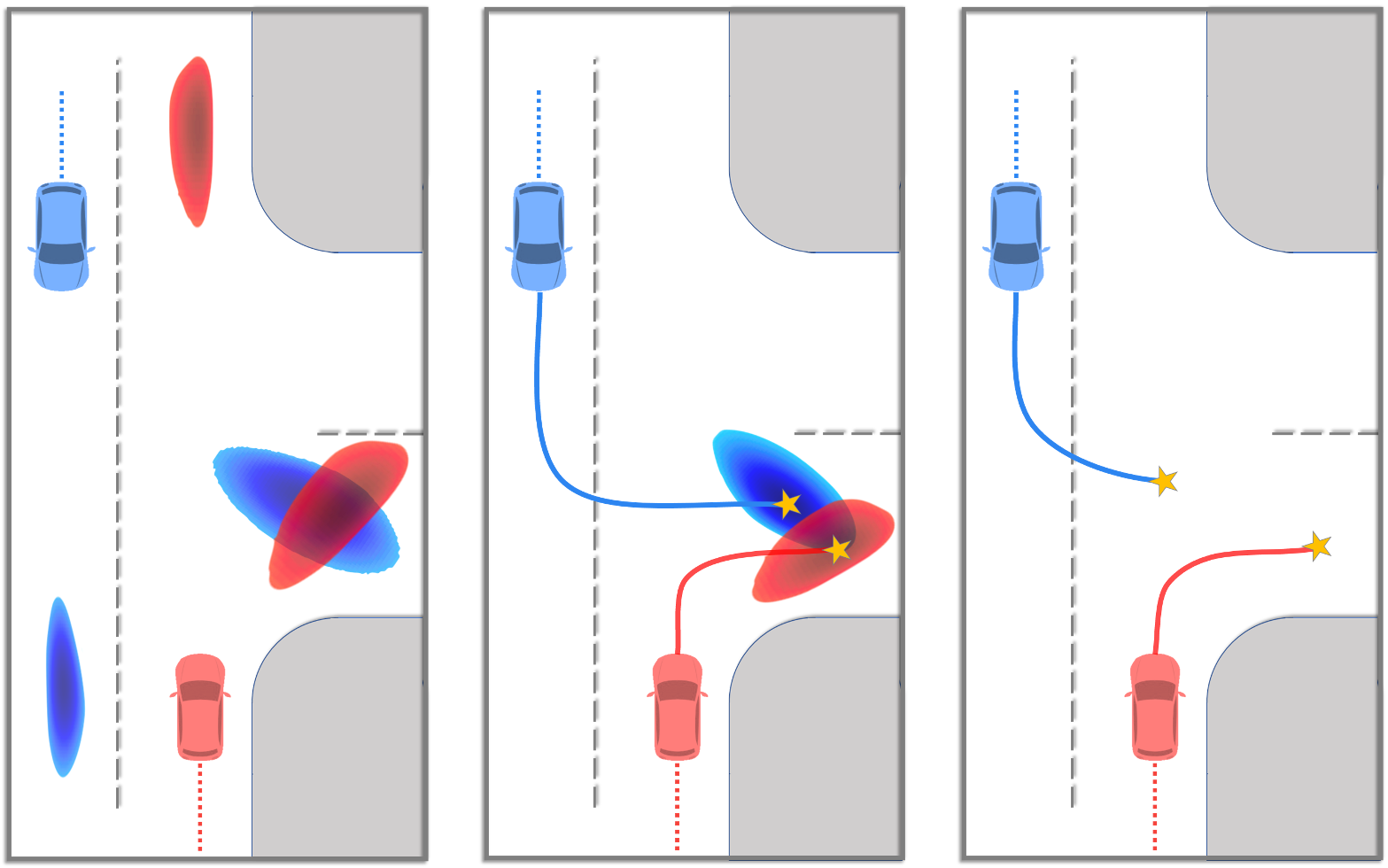}
    \caption{A motivating example of BiFF. \textbf{Left}: Marginal heatmaps of two interactive agents conflict with each other. \textbf{Middle}: For each scene modality, assignment scores generated by the prediction header become decoupled by fusing high-level future intentions across agents. \textbf{Right}: For each scene modality, the predicted trajectories are more scene-consistent by fusing low-level future behaviors across agents.}
\vspace{-3mm}

\label{fig:illustration}
\end{figure}

Most existing studies on trajectory prediction \cite{chai2019multipath, gao2020vectornet, gu2021densetnt, lee2017desire, liang2020learning} tend to focus on generating marginal prediction samples of future trajectories for each agent, without considering their future interactions. Thus, such marginal prediction models may generate trajectories with a high overlap rate \cite{sun2022m2i}, then the unreliable results are provided for the downstream planning module. To overcome this limitation, recent works \cite{gilles2021thomas, girgis2022latent, ngiam2022scene} propose joint motion prediction models that generate scene-compliant trajectories. However, previous models focus on capturing interaction across agents in the tracking history, explicitly modeling future interaction remains an open problem. The preliminary research regarding future interaction is conducted in the recent two works. M2I \cite{sun2022m2i} classifies interactive agents as influencer and reactor. Conditioning on the influencer, the reactor benefit from future information, but the influencer is not optimized. MTR \cite{shi2022motion} predicts future trajectories for all the surrounding agents, but the future information is not reliable enough since trajectories are directly regressed using polyline features without any iterative refinement. To model the future interactions, we propose Bi-level Future Fusion (BiFF), and the motivating example is presented in Fig. \ref{fig:illustration}, in which the unrealistic conflicting prediction between the marginal heatmaps is mitigated by the proposed High-level Future Intentions Fusion (HFIF) and Low-level Future Behaviors Fusion (LFBF).

Moreover, regarding these models \cite{gilles2021thomas, girgis2022latent, ngiam2022scene}, all target agents are normalized to scene-centric coordinate, which sacrifices prediction accuracy and model generalization, therefore, requires heavy labor in data augmentation. One possible solution is to simply extract local features for agent-centric representation \cite{zhou2022hivt}, but it is not memory-efficient with redundant context encoding between different target agents. 
To overcome these shortcomings, polyline-based coordinate is specifically designed to make all predictions invariant to the global reference frame, and the feature fusion between different coordinates is performed with relative positional encoding.

To summarize, our contributions remain as follows: (1) We propose a novel Bi-level Future Fusion (BiFF) model that incorporates a High-level Future Intention Fusion (HFIF) mechanism to generate scene-consistent goals and a Low-level Future Behavior Fusion (LFBF) mechanism to predict scene-compliant trajectories. (2) For multi-agent prediction, we design polyline-based coordinates to provide agent-centric representations for all target agents without redundant context encoding, which is memory-saving, data-efficient, and robust to the variance of global reference frame. (3) Our approach achieves state-of-the-art performance on the interactive prediction benchmark of Waymo Open Motion Dataset (WOMD).

\setcounter{figure}{1}  
\begin{figure*}[t]
\centering
\includegraphics[width=1.0\linewidth]{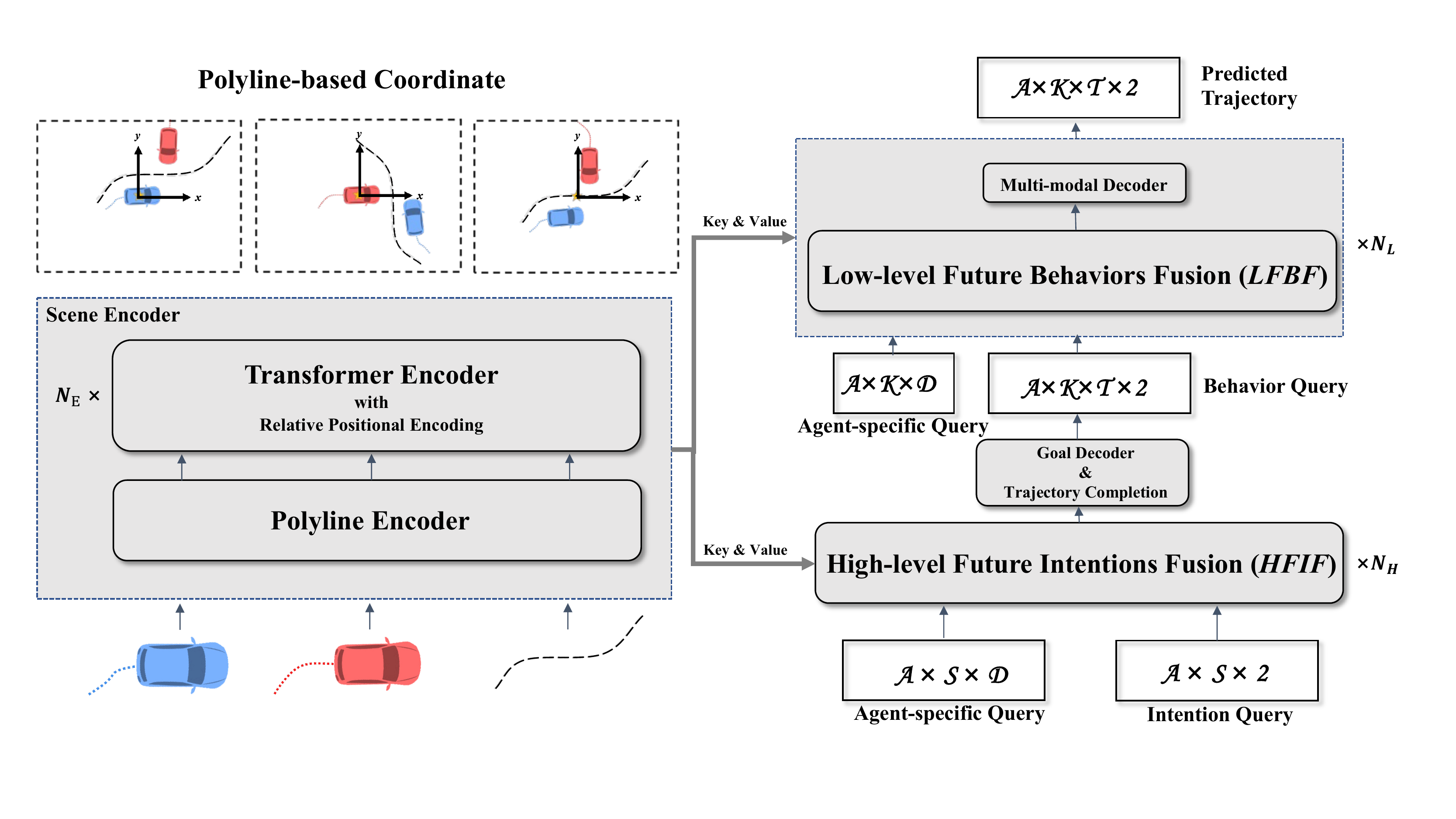}
\vspace{-7mm}
\caption{  BiFF framework overview. Left indicates the scene context encoder, with agents (in red and blue) and road (in dashed curves) normalized to their own polyline-based coordinate separately. Right shows the motion decoder, with $A$ predicted interactive agents, $S$ static intentions (conditional anchors), $D$ hidden feature dimension, $K$ predicted scene modalities, and $T$ predicted future steps. $N_E$, $N_L$ and $N_H$ are the number of stacked layers. Details can be referred to Model Overview (Sec.\ref{Model-Overview}).
}
\label{fig:framework}

\end{figure*}

\section{Related Work}
\label{sec:Related-Work}

Recently, the rise in autonomous driving, as well as the availability of the corresponding datasets and benchmarks \cite{caesar2020nuscenes, ettinger2021large, zhan2019interaction}, have prompted significant research interest in motion prediction. Due to the uncertainty in human intent, numerous approaches have been explored to model the intrinsic multi-modal distribution. Earlier works employ stochastic models such as generative adversarial networks (GANS) \cite{gupta2018sgan, sadeghian2019sophie, zhao2019multi} or conditional variational autoencoders (CVAES) \cite{casas2020implicit, lee2017desire, rhinehart2018r2p2, tang2019multiple} to draw samples approximating the output distribution. However, sample inefficiency at inference time prohibits them from being deployed in safety-critical driving scenarios. Alternatively, some works, including \cite{chai2019multipath, hong2019rules, mercat2020multi, salzmann2020trajectron++, song2020pip}, utilize Gaussian Mixture Models (GMMs) to parameterize multi-modal predictions and generate compact distributions. More recent work conduct classification on a set of predefined anchor trajectories \cite{chai2019multipath, phan2020covernet, song2022learning} or potential intents such as goal targets \cite{rhinehart2019precog, gilles2021home, gu2021densetnt, zhao2020tnt}, such models have shown remarkable success in popular motion forecasting benchmark.

\subsection{Scene Context Representation}
\label{sec:scene}

Motion prediction models typically take road map information and agent history states as input. To encode such scene context efficiently, earlier works \cite{alahi2016slstm, cui2019multimodal,  deo2018cslstm,djuric2020uncertainty, gupta2018sgan, phan2020covernet} rasterize them into multi-channel BEV images for processing with convolutional neural networks (CNNs). Such models predominantly operate at a scene level and sacrifice the pose-invariance of each target. In addition, recent studies have demonstrated the efficiency and scalability of encoding scene context using vectorized data directly from high-definition (HD) maps, as evidenced by the success of \cite{gao2020vectornet, liang2020learning, ye2021tpcn, gilles2022gohome, varadarajan2022multipath++}. All targets still share the global frame in the vectorized methods above.
By utilizing relative positional encoding, the agent-centric schemes \cite{zhou2022hivt, cui2022gorela, jia2022multi} highly related to us are proposed, as they learn context representations that are invariant to the global transformation. However, our proposed method differs in three significant ways. (1) The redundant map information in HiVT \cite{zhou2022hivt} is precluded. (2) The feature encoder in our polyline-based coordinate is more straightforward than \cite{zhou2022hivt, cui2022gorela, jia2022multi}, where graph nodes aggregates features with hand-crafted design. (3) We incorporate future interactions in the agent-coordinate frame, similar to scene context encoding.

\subsection{Interactive Trajectory Prediction}
\label{sec:Interactive}

Multi-agent motion prediction is a crucial area, and one of the most challenging problems in this field is predicting scene-compliant future trajectories. To address this problem, agent interactions have been modelled using various techniques, such as hand-crafted functions \cite{helbing1995social}, graph neural networks \cite{liang2020learning, mohamed2020social, li2019grip, li2020evolvegraph} and attention mechanisms \cite{song2022learning, gao2020vectornet, mercat2020multi, gilles2021thomas, chen2022intention}. Recent studies have focused on using Transformers \cite{vaswani2017attention} to capture the interactions among agents and map elements \cite{ngiam2022scene, girgis2022latent, salzmann2020trajectron++,  liu2021multimodal, yuan2021agentformer}. While these models excel in modeling interactions in the encoding part, they tend to independently predict the final trajectory candidates losing scene-compliance. 

In order to generate scene-consistent outputs, ILVM \cite{casas2020implicit} uses scene latent variables for interaction modeling and optimizes with ELBO. SceneTransformer \cite{ngiam2022scene} and AutoBots \cite{girgis2022latent} decode joint motion predictions with learnable queries in the Transformer block. Thomas \cite{gilles2021thomas} directly designed a recombination model to recombine the agents' endpoints generating joint futures. M2I \cite{sun2022m2i} identifies an influencer-reactor pair. To explicitly capture future interaction, the trajectory of reactor is predicted depending on influencer. JFP \cite{luo2022jfp}  utilizes a dynamic interactive graph model on top of the trajectory candidates to generate consistent future predictions. MTR \cite{shi2022motion} presents an auxiliary dense future prediction task to facilitate future interaction modelling and guide the model to generate scene-compliant trajectories.

\section{Approach}
\label{Approach}

\subsection{Model Overview}
\label{Model-Overview}

The overall framework of BiFF is illustrated in \figref{fig:framework}. For the scene encoder, we represent the driving scene in a vectorized format for both agents and roads normalized to their own polyline-based coordinate separately, which achieves agent-centric representation for predicted agents without any redundant context encoding (Sec.\ref{sec:Scene-Encoder}). Then agents and roads are encoded by a PointNet-like \cite{qi2017pointnet} polyline encoder followed by transformer encoder with relative positional encoding. For the motion decoder, the High-level Future Intentions Fusion (HFIF) module fuses static future intentions across target agents followed by multi-modal goal decoder and trajectory completion (Sec.\ref{sec:HFIF}). Given the predicted trajectory from HFIF, we take the Low-level Future Behaviors Fusion (LFBF) module to fuse dynamic future behaviors of interactive agents in each scene modality (Sec.\ref{sec:LFBF}). Finally, we present the details of training loss and model inference (Sec.\ref{sec:Training-Loss}).

\subsection{Scene Encoder with Polyline-based Coordinate}
\label{sec:Scene-Encoder}

For the joint trajectory prediction, existing works apply scene-centric representation normalizing all context to a unique global frame
\cite{gilles2021thomas, girgis2022latent, ngiam2022scene}. We argue that polyline-based coordinate with relative positional encoding is more efficient for interactive prediction. 


{\bf Represent context with polyline-based coordinate.} We adopt vectorized method proposed in \cite{gao2020vectornet} to represent agents and roads as polylines. The agent-centric strategy is utilized with polyline-based coordinate. 
For the pose normalization, each agent is centered to their own pose at the current time step while each road is normalized to the center of its own polyline. With the normalized features, for the polyline construction, each agent is composed of all history frames with state features (e.g. position, velocity, heading, type, etc) while each road consists of maximum 10 points sampled at intervals of 1 meter with point information (e.g., position, direction, type, etc.). Finally, we use PointNet-like \cite{qi2017pointnet} architecture with an MLP followed by max pooling per polyline to obtain the feature per element of agent or road.   

{\bf Fuse polylines with relative positional encoding.} We leverage transformer encoder stacked with $N_E$ layers to fuse features across polylines. Considering that all polylines are self-normalized, to make each polyline aware of the poses of others, the relative positional encoding is utilized for performing pair-wise message passing. We take the directional feature fusion between polylines $\mathbf{p}_i$ and  $\mathbf{p}_j$ (i.e., $j \to i$) as an example, where the pairwise relative positional encoding $\mathbf{p}_{ij}$ is obtained using MLP to the relative pose (e.g., ${\Delta}x_{ij}, {\Delta}y_{ij}, \text{cos}({\Delta}{\theta}_{ij}), \text{sin}({\Delta}{\theta}_{ij})$). Concretely, by normalizing the coordinate of polyline $j$ to polyline $i$, (${\Delta}x_{ij}$, ${\Delta}y_{ij}$) is the origin location and ${\Delta}{\theta}_{ij}$ is the rotation angle of polyline $j$ in the coordinate of polyline $i$. Then we incorporate $\mathbf{p}_{ij}$ into the vector transformation. The polyline $i$ is used to calculate query vector, and the transformed polyline $j$ is considered as key and value vectors:







\begin{equation}
\begin{gathered} 
  \mathbf{q}_i^e = \mathbf{W}_q^{e} \mathbf{h}_i^{e}, \\
  \mathbf{k}_{ij}^e = \mathbf{W}_k^{e} \mathbf{h}_j^{e} + \mathbf{W}_{pos}^{e} \mathbf{p}_{ij}, \\ \mathbf{v}_{ij}^e = \mathbf{W}_v^{e} \mathbf{h}_j^{e}
\end{gathered}
  \label{eq:important}
\end{equation}

where $\mathbf{h}_i^{e}$ and $\mathbf{h}_j^{e}$ are the features of polyline $i$ and polyline $j$ respectively, 
$\mathbf{W}_q^{e}, \mathbf{W}_k^{e}, \mathbf{W}_v^{e}, \mathbf{W}_{pos}^{e} \in \mathbb{R}^{d \times d}$ are learnable matrices for linear projection, and $d$ is the dimension of polyline feature. The weight matrices are not shared across different encoder layers. 
The scaled dot-product attention is calculated by the given query, key and value:
\begin{equation}
{\alpha}_{ij}^e=\frac{\operatorname{exp}\left ( \mathbf{q}_{i}^{e^{\top}} \mathbf{k}_{i j}^{e}/ \sqrt {d_k} \right )}{\sum_{j \in \mathcal{N}_{i}}\operatorname{exp} \left ( \mathbf{q}_{i}^{e^{\top}} \mathbf{k}_{i j}^{e}/ \sqrt {d_k} \right ) }, 
  \label{eq:compute_weight}
\end{equation}
\begin{equation}
  \mathbf{\hat{h}}_{i}^{e}=\sum_{j \in \mathcal{N}_{i}} \mathbf{\alpha}_{i j}^{e} \mathbf{v}_{i j}^{e}, 
  \label{eq:sum_weight}
\end{equation}

where $\mathcal{N}_{i}$ is the set of polyline $i$'s neighbors with $k$ closest polylines (whose polyline coordinates are closest to the query polyline $i$), $d_k$ is the dimension of query or key. The updated feature $\mathbf{\hat{h}}_{i}^{e}$ output from the final encoder layer is considered as polyline feature for the motion decoder. For simplicity, we omit the commonly used Multi-Head Attention, Layer Normalization, Feed-Forward Network and Residual Connection of Transformer \cite{vaswani2017attention} in the equations.

\subsection{High-level Future Intentions Fusion (HFIF)}
\label{sec:HFIF}
\begin{figure*}[ht]
\centering
\includegraphics[width=1.0\linewidth]{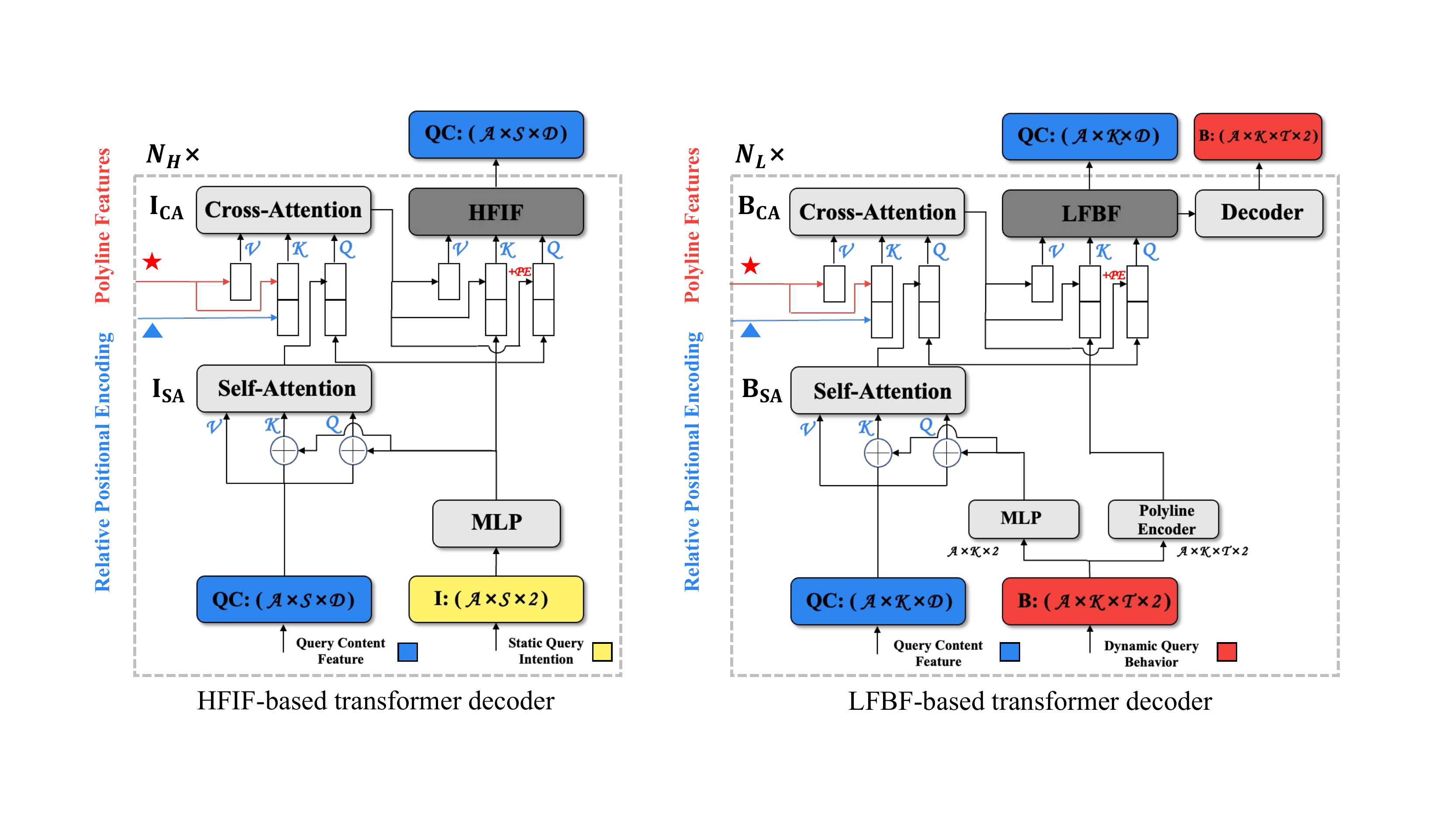}
\vspace{-12mm}
\caption{ Details of HFIF-based and LFBF-based decoder network structure.
}
\label{fig:decoder}
\vspace{-5mm}
\end{figure*}

Given the marginal target heatmap of each predicted agent \footnote {The details of model architecture and training for predicting marginal heatmap can be found in Appendix.}, we select top $S$ intentions with the highest scores as conditional anchors for High-level Future Intentions Fusion (HFIF). As shown in the left part of Fig. \ref{fig:decoder}, the transformer decoder with static query intentions is utilized to localize each goal and aggregate intention-specific context. To fuse the high-level future intentions, the decoder is further extended with HFIF. Finally, $K$ group goals are generated with multi-modal decoder followed by a simple trajectory completer. Note that in both HFIF (Sec.\ref{sec:HFIF}) and LFBF (Sec.\ref{sec:LFBF}), all features fusion between different coordinates are performed with relative positional encoding, and the learnable weights are not shared across different decoder layers. The HFIF is stacked with $N_H$ layers, and the detailed architecture is illustrated in the following. 

{\bf Intention queries and attention module.} In the HFIF-based transformer decoder layer, to distinguish queries of different agents, we initialized them using the fused feature $\mathbf{h}^{e}$ of corresponding target agent obtained from the scene encoder (Sec.\ref{sec:Scene-Encoder}) in the first decoder layer, but using the query content feature output from previous layer in the following decoder layers. Then the agent-specific queries are further augmented with the corresponding static query intentions in all decoder layers. Given $S$ intentions normalized to the predicted agent's polyline coordinate, we apply MLP to encode each query intention $\mathbf{s} \in \mathbb{R}^2$ (x, y) as $\mathbf{e}^{qi} \in \mathbb{R}^d$ and add it to the query content. Then the self-attention layer ($\bm{\mathrm{I}_\mathrm{SA}}$) is utilized to propagate information among $S$ queries inner each agent, and the updated intention query $\mathbf{h}^{isa} \in \mathbb{R}^d$ is obtained. 

Next, we introduce cross-attention block ($\bm{\mathrm{I}_\mathrm{CA}}$) with relative positional encoding to aggregate intention-specific context. For example, any intention of any target agent $i$ is used as the query vector, and any polyline $j$ is considered as key and value vectors. Motivated by \cite{shi2022motion}, to decouple the attention contribution from two types of features, we concatenate query content with intention embedding while concatenate key content with positional encoding:



\begin{equation}
\begin{gathered} 
  \mathbf{q}_i^{ica} = \left [ \mathbf{W}_q^{ica} \mathbf{h}_i^{isa}, \mathbf{W}_{qe}^{ica} \mathbf{e}_i^{qi} \right ] , \\
  \mathbf{k}_{ij}^{ica} = \left [ \mathbf{W}_k^{ica} \mathbf{h}_j^{e},\mathbf{W}_{pos}^{ica} \mathbf{p}_{ij} \right ], \\ \mathbf{v}_{ij}^{ica} = \mathbf{W}_v^{ica} \mathbf{h}_j^{e},
\end{gathered}
  \label{eq:hfif_hca}
\end{equation}


where the intention feature $\mathbf{h}_i^{isa}$ is returned by self-attention module paired with the intention embedding $\mathbf{e}_i^{qi}$, the polyline feature $\mathbf{h}_j^{e}$ is output from scene encoder, and $\mathbf{p}_{ij}$ is the relative positional encoding (normalize polyline coordinate of polyline $j$ to target agent $i$). All $\left [ \cdot, \cdot \right ]$ denotes concatenating two vectors $\in \mathbb{R}^d$ as one vector $\in \mathbb{R}^{2d}$ and all $\mathbf{W}$ (omit superscript and subscript) are learnable matrices from Eq. \ref{eq:hfif_hca} to Eq. \ref{eq:lfbf_l}. Considering that information aggregation from agent or road is inherently distinguished in the attention mechanism, we adopt two groups of projection matrices for agent and road separately in both HFIF (Sec.\ref{sec:HFIF}) and LFBF (Sec.\ref{sec:LFBF}). 
By calculating cross attention using Eq. \ref{eq:compute_weight} and Eq. \ref{eq:sum_weight}, for any intention of any target agent $i$, the updated query intention $\mathbf{h}_i^{ica} \in \mathbb{R}^d$ is obtained.  
The $\mathcal{N}_{i}$ in Eq. \ref{eq:compute_weight} is composed of features of all agent polylines and the closest $L$ road polylines (whose centers are closest to the current query intention embedded as $\mathbf{e}_i^{qi}$) output from scene encoder.

\begin{figure}[ht]
    \centering
    \includegraphics[width=0.8\columnwidth]{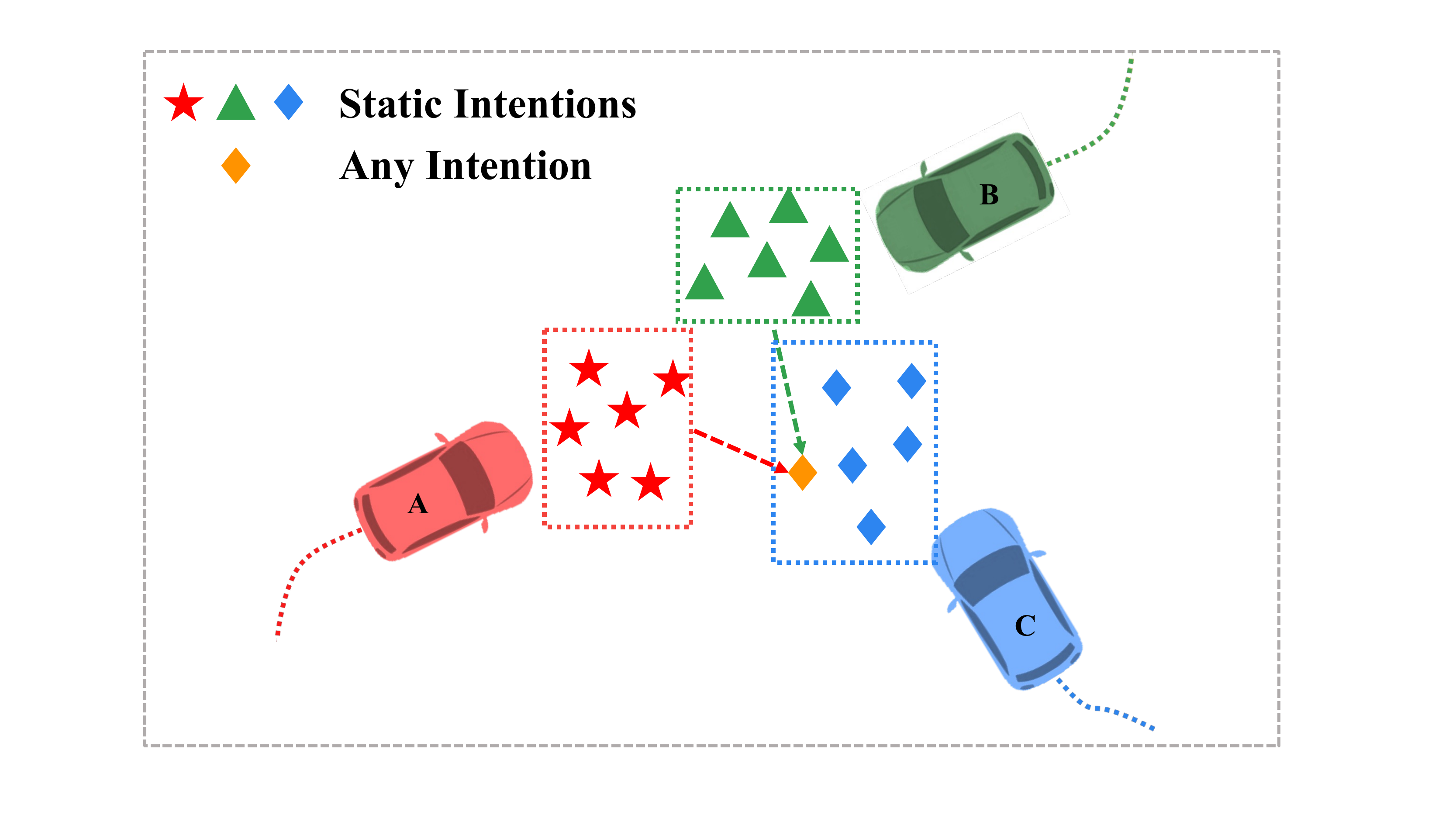}
    \caption{HFIF: any intention aggregates features of all static intentions from other target agents.}
    \vspace{-4mm}
\label{fig:static_intentions}
\end{figure}


{\bf Fuse high-level future intentions across agents.} 
We introduce \textbf{HFIF} module to capture the interaction of future intentions across target agents. Fig. \ref{fig:static_intentions} shows the details of HFIF, where any intention of any target agent aggregates features of all static intentions from other targets using cross-attention. For example, when performing message passing from any intention of target agent $j$ to any intention of target agent $i$, the fused intention features paired with their original embedding are incorporated into the attention mechanism with relative positional encoding $\mathbf{p}_{ij}$ (normalize polyline coordinate of target agent $j$ to target agent $i$):

\begin{equation}
\begin{gathered} 
  \mathbf{q}_i^{hif} = \left [ \mathbf{W}_q^{hif} \mathbf{h}_i^{ica}, \mathbf{W}_{qe}^{hif} \mathbf{e}_i^{qi} \right ], \\
  \mathbf{k}_{ij}^{hif} = \left [ \mathbf{W}_k^{hif} \mathbf{h}_j^{ica} + \mathbf{W}_{pos}^{hif} \mathbf{p}_{ij}, \mathbf{W}_{ke}^{hif} \mathbf{e}_j^{qi} \right ], \\ \mathbf{v}_{ij}^{hif} = \mathbf{W}_v^{hif} \mathbf{h}_j^{ica},
\end{gathered}
  \label{eq:hfif_h}
\end{equation}


where $\mathbf{h}_i^{ica}$ and $\mathbf{h}_j^{ica}$ output from cross attention module is the scene-based features of any intention of target agent $i$ and agent $j$, while $\mathbf{e}_i^{qi}$, $\mathbf{e}_j^{qi}$ are the corresponding intention embedding, respectively. To make $\mathbf{e}_i^{qi}$ aware of the position of $\mathbf{e}_j^{qi}$, the location of any intention of agent $j$ is normalized to the polyline coordinate of agent $i$ before embedding. The computation in Eq. \ref{eq:compute_weight} and Eq. \ref{eq:sum_weight} is conducted for the high-level intention fusion ($hif$) with the neighbor $\mathcal{N}_{i}$ containing all intention features of other target agents. Finally, we obtain the updated feature $\mathbf{h}_i^{hif} \in \mathbb{R}^d$ for any intention of any target agent $i$, and the feature is used as the initialized query content for the next decoder layer. 

\begin{figure}[ht]
    \centering
    \includegraphics[width=0.8\columnwidth]{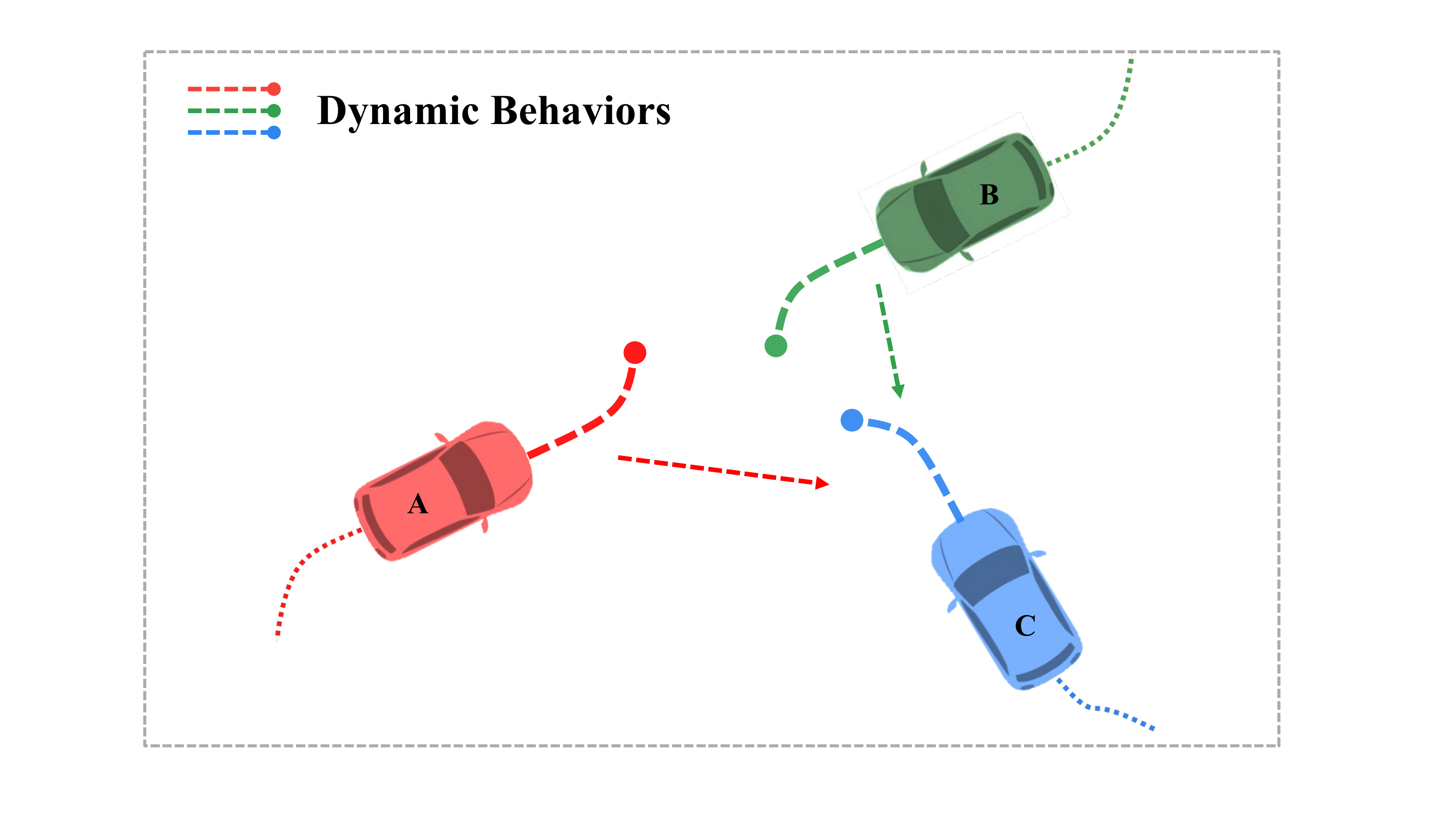}
    \caption{LFBF: any behavior aggregates features of all dynamic behaviors in the same scene modality from other target agents.}
    \vspace{-2mm}
\label{fig:dynamic_behaviors}
\end{figure}

{\bf Multi-modal decoder in HFIF.} Given $\mathbf{h}_i^{hif}$ output from the final decoder layer in HFIF, we introduce the multi-modal decoder composed of goal regression followed by trajectory completion. To extract different interactive modes, we apply a multi-modal goal decoder containing $K = 6$ headers to the fused intention features from HFIF-based transformer decoder, and all headers are consist of 1D convolution with identical structure. Each header predicts scene-consistent goals for all target agents in each scene modality. For example, to regress goals of the $a$-th agent in the $k$-th modality, all fused intention features of the $a$-th agent are passed through the $k$-th header followed by a Softmax function to predict the assignment scores ${\gamma}^a_{k,s}$ ($s$ is the $s$-th intention in all $S$ conditional anchors). The corresponding regressing goal $\hat{g}^{a}_k$ can be computed by the weighted sum of the conditional anchors' coordinate $p^a_s$, i.e., $\hat{g}^{a}_k= {\textstyle \sum_{s=1}^{S}} {\gamma}^a_{k,s} p^a_s $. Given the goals predicted for each scene modality, we concatenate each goal with the corresponding agent feature from scene encoder (Sec.\ref{sec:Scene-Encoder}), then pass the new embedding to an MLP to complete predicted trajectory for each target in each scene modality.

\subsection{Low-level Future Behaviors Fusion (LFBF)}
\label{sec:LFBF}

After HFIF, we obtain $K$ groups of predicted trajectories considered as dynamic anchors for Low-level Future Behaviors Fusion (LFBF). Refer to the right part of Fig. \ref{fig:decoder}, the transformer decoder with dynamic query behaviors (predicted trajectories) is utilized to localize each modality and aggregate behavior-specific context. We extend decoder with LFBF to fuse the low-level future behaviors in each scene modality. The LFBF-based transformer decoder is stacked with $N_L$ layers to iteratively refine predicted trajectories and fuse future behaviors. The details are illustrated as follows. 

{\bf Behavior queries and attention module.} In the LFBF-based transformer decoder layer, the dynamic query behaviors are initialized by the trajectories generated from HFIF, and dynamically updated with the trajectories predicted from previous decoder layer, which is designed for iteratively trajectory refine \cite{shi2022motion}. Given any future trajectory $\mathbf{y}^{1:T} \in \mathbb{R}^{T \times 2}$ normalized to the corresponding polyline coordinate of any target agent, we encode the query behavior as $\mathbf{e}^{qb} \in \mathbb{R}^d$ using polyline encoder (a small PointNet-like \cite{qi2017pointnet} structure) and encode the endpoint $\mathbf{y}^T \in \mathbb{R}^2$ as $\mathbf{e}^{qeb} \in \mathbb{R}^d$ using MLP. The $\mathbf{e}^{qb}$ is applied to the scene context aggregation and future behaviors fusion while $\mathbf{e}^{qeb}$ is utilized to localize each modality. Regarding initializing query content in each decoder, we take the same process in HFIF, and the queries are added with dynamic $\mathbf{e}^{qeb}$ to distinguish each modality in all decoder layers. For the self-attention layer ($\bm{\mathrm{B}_\mathrm{SA}}$), we apply it to propagate information among $K$ modalities inner each agent, and the updated behavior query $\mathbf{h}^{bsa} \in \mathbb{R}^d$ is obtained. 

Next, we introduce cross-attention block ($\bm{\mathrm{B}_\mathrm{CA}}$) with relative positional encoding to aggregate behavior-specific scene context. For example, any future behavior of any target agent $i$ is used as the query vector, and any polyline $j$ is considered as key and value vectors:



\begin{equation}
\begin{gathered} 
  \mathbf{q}_i^{bca} = \left [ \mathbf{W}_q^{bca} \mathbf{h}_i^{bsa}, \mathbf{W}_{qe}^{bca} \mathbf{e}_i^{qb} \right ] , \\
  \mathbf{k}_{ij}^{bca} = \left [ \mathbf{W}_k^{bca} \mathbf{h}_j^{e},\mathbf{W}_{pos}^{bca} \mathbf{p}_{ij} \right ], \\ \mathbf{v}_{ij}^{bca} = \mathbf{W}_v^{bca} \mathbf{h}_j^{e},
\end{gathered}
  \label{eq:lfbf_lca}
\end{equation}


where the behavior feature $\mathbf{h}_i^{bsa}$ is returned by self-attention module paired with the behavior embedding $\mathbf{e}_i^{qb}$, polyline feature $\mathbf{h}_j^{e}$ is output from scene encoder, and $\mathbf{p}_{ij}$ is the relative positional encoding (normalize coordinate of polyline $j$ to target agent $i$). For any target agent $i$ in any scene modality, we update the query behavior as $\mathbf{h}_i^{bca} \in \mathbb{R}^d$ by calculating cross attention with Eq. \ref{eq:compute_weight} and Eq. \ref{eq:sum_weight}, where $\mathcal{N}_{i}$ is composed of features of all agent polylines and the closest $L$ road polylines (whose centers are closest to the center of the current query behavior embedded as $\mathbf{e}_i^{qb}$).

{\bf Fuse low-level future behaviors across agents.} In each scene modality, we introduce \textbf{LFBF} module to capture the interaction of low-level future behaviors across target agents. Fig. \ref{fig:dynamic_behaviors} illustrates the details of LFBF, where any behavior of any target agent aggregates features of all dynamic behaviors in the same scene modality from other targets using cross-attention. For example, when performing message passing from future behavior of target agent $j$ to target agent $i$ in any scene modality, the fused future behavior features paired with their trajectory embedding are introduced to calculate query, key and value vectors with relative positional encoding $\mathbf{p}_{ij}$ (normalize polyline coordinate of target agent $j$ to target agent $i$):

\begin{equation}
\begin{gathered} 
  \mathbf{q}_i^{lbf} = \left [ \mathbf{W}_q^{lbf} \mathbf{h}_i^{bca}, \mathbf{W}_{qe}^{lbf} \mathbf{e}_i^{qb} \right ], \\
  \mathbf{k}_{ij}^{lbf} = \left [ \mathbf{W}_k^{lbf} \mathbf{h}_j^{bca} + \mathbf{W}_{pos}^{lbf} \mathbf{p}_{ij}, \mathbf{W}_{ke}^{lbf} \mathbf{e}_j^{qb} \right ], \\ \mathbf{v}_{ij}^{lbf} = \mathbf{W}_v^{lbf} \mathbf{h}_j^{bca},
\end{gathered}
  \label{eq:lfbf_l}
\end{equation}


where $\mathbf{h}_i^{bca}$ and $\mathbf{h}_j^{bca}$ are the features of future behaviors output from cross attention module for any target agent $i$ and agent $j$ in any scene modality, while $\mathbf{e}_i^{qb}$, $\mathbf{e}_j^{qb}$ are the corresponding behavior embedding, respectively. To make $\mathbf{e}_i^{qb}$ aware of the position of $\mathbf{e}_j^{qb}$, the location of any future trajectory of agent $j$ is normalized to the polyline coordinate of agent $i$ before embedding. The resulting query, key and value are utilized for low-level behavior fusion ($lbf$) using Eq. \ref{eq:compute_weight} and Eq. \ref{eq:sum_weight}, where the neighbor $\mathcal{N}_{i}$ contains all future behavior features in the current scene modality from other target agents. Finally, we obtain the updated feature $\mathbf{h}_i^{lbf} \in \mathbb{R}^d$ for the future behavior of any target agent $i$ in any scene modality, and the feature is used as the initialized query content for the next decoder layer. 

{\bf Multi-modal decoder in LFBF.} For each LFBF-based decoder layer, we append an MLP block to the output feature $\mathbf{h}_i^{lbf}$ for predicting future trajectory of any target agent $i$ in any scene modality. 

\subsection{Training Loss}
\label{sec:Training-Loss}

The training loss is composed of goal regression $L_{G}$ and trajectory regression $L_{T}$ for the total $A$ target agents. For the multiple scene modalities, a winner-takes-all (WTA) strategy is taken to address mode collapse. More specifically, both two losses are only calculated for the scene modality that has the minimum final displacement error summed by all target agents, e.g., the sum of Euclidean distance between the endpoint of predicted trajectory (output from the final decoder layer in LFBF) and ground truth: 
\begin{equation}
  L_{G} = \sum_{a=1}^{A} L_{reg}(g^a, \hat{g}^a_{k^*} ), \quad L_{T} = \sum_{a=1}^{A} L_{reg}({\bf y}^a, \hat{\bf y}^a_{k^*} )
  \label{eq:important}
\end{equation}

where $g^a$ and $\hat{g}^a_{k^*}$ are the GT endpoint and regressed goal from HFIF decoder, respectively. ${\bf y}^a$ and $\hat{\bf y}^a_{k^*}$ are the GT trajectory and predicted trajectory from LFBF-based decoder layer, respectively. $L_{reg}$ is the smooth $l1$ loss. 
$k^*$ is the index of the selected scene modality. The final loss is the sum of the $L_G$ and the $L_T$ with equal weights, where the $L_T$ is applied to each LFBF-based decoder layer.

For inference, trajectories generated by the last decoder layer are considered as the final prediction result. The joint likelihood of each scene modality is calculated by the product of all target agent scores in each modality. To obtain agent modality score, we take the endpoint of each predicted trajectory to query the closest intention in the corresponding marginal heatmap (conditional anchors), then the queried intention score is used to calculate joint likelihoods.


\begin{table*}[ht]
\definecolor{Gray}{gray}{0.90}
\definecolor{themeblue}{RGB}{57, 162, 219}
\newcolumntype{Z}{S[table-format=2.2,table-auto-round]}
\newcolumntype{g}{>{\columncolor{Gray}}S[table-format=2.2,table-auto-round]}
\centering
\setlength{\tabcolsep}{1.0mm}
\ra{1.1}
\footnotesize
\caption{
Comparison with top-ranked entries on the \href{https://waymo.com/open/challenges/2021/interaction-prediction/}{WOMD Interaction Leaderboard}. ${\dagger}$ refer to ensemble methods.
}
\begin{tabular}{@{}cZ@{}l@{}cZZZcZZZcZZZ@{}}
  \toprule
  
  && \multirow{2}[3]{*}{Method} && \multicolumn{3}{c}{Vehicle(8s)} 
  && \multicolumn{3}{c}{Pedestrain(8s)} && \multicolumn{3}{c}{Cyclist(8s)}\\
  
  \cmidrule(l{1mm}r{1mm}){5-7} \cmidrule(l{1mm}r{1mm}){9-11} \cmidrule(l{1mm}r{1mm}){13-15}
  &&&&{\scriptsize \quad mFDE$\downarrow$} & {\scriptsize \quad MR$\downarrow$ } & {\scriptsize \quad mAP$\uparrow$ }&& 
  {\scriptsize \quad mFDE$\downarrow$} & {\scriptsize \quad MR$\downarrow$ } & {\scriptsize \quad mAP$\uparrow$ }&&
  {\scriptsize \quad mFDE$\downarrow$} & {\scriptsize \quad MR$\downarrow$ } & {\scriptsize \quad mAP$\uparrow$ } \\

  \midrule
  \multirow{9}[3]{*}{Test}
  && Waymo LSTM baseline \cite{ettinger2021large} 
  && 12.40 & 0.87  & 0.01 
  && 6.85 & 0.92  & 0.00  
  && 10.84 & 0.97  & 0.00 
  \\
  && HeatIRm4 \cite{mo2021multi}
  && 7.20 & 0.80  & 0.07 
  && 4.06 & 0.80  & 0.05  
  && 6.69 & 0.85  & 0.01  
  \\
  && AIR$^{2}$ \cite{wu2021air}
  && 5.00 & 0.64  & 0.10 
  && 3.68 & 0.71  & 0.04  
  && 5.47 & 0.81  & 0.04  
  \\
  && SceneTransformer \cite{ngiam2022scene}
  && 4.08 & 0.50  & 0.10 
  && 3.19 & 0.62  & 0.05  
  && 4.65 & 0.70  & 0.04  
  \\
  && DenseTNT \cite{gu2021densetnt}
  && 4.75 & 0.52  & \bfseries 0.20 
  && 3.32 & 0.59  & \bfseries 0.11  
  && 5.15 & 0.73  & \bfseries 0.05  
  \\
  && M2I \cite{sun2022m2i}
  && 5.65 & 0.57  & 0.16 
  && 3.73 & 0.60  & 0.06  
  && 6.16 & 0.74  & 0.03  
  \\
  &&\cellcolor{gray!40}BiFF (Ours)
  &&\cellcolor{gray!40} 3.82 &\cellcolor{gray!40} \bfseries 0.46  &\cellcolor{gray!40} 0.13 
  &\cellcolor{gray!40}&\cellcolor{gray!40} 2.79 &\cellcolor{gray!40} \bfseries 0.54  &\cellcolor{gray!40} 0.06  
  &\cellcolor{gray!40}&\cellcolor{gray!40} 4.33 &\cellcolor{gray!40} \bfseries 0.68  &\cellcolor{gray!40} 0.03  
  \\
  &&\cellcolor{gray!40}BiFF-200 (Ours)
  &&\cellcolor{gray!40} \bfseries 3.71 &\cellcolor{gray!40} 0.47  &\cellcolor{gray!40} 0.12 
  &\cellcolor{gray!40}&\cellcolor{gray!40} \bfseries 2.73 &\cellcolor{gray!40} 0.56  &\cellcolor{gray!40} 0.05  
  &\cellcolor{gray!40}&\cellcolor{gray!40} \bfseries 4.29 &\cellcolor{gray!40} 0.69  &\cellcolor{gray!40} 0.03  
  \\
  \cmidrule(l{0mm}r{0mm}){3-15}
  &&  ${\dagger}$ MTR \cite{shi2022motion}
  && 4.04 & 0.49  & 0.21 
  && 2.86 & 0.53  & 0.13  
  && 4.24 & 0.65  & 0.05 
  \\
  && ${\dagger}$ JFP \cite{luo2022jfp}
  && 3.88 & 0.45  & 0.19 
  && 2.81 & 0.54  & 0.15  
  && 4.19 & 0.63  & 0.06 
  \\
  \midrule
  \midrule
  \multirow{2}[3]{*}{Val}
  && Waymo Full Baseline \cite{ettinger2021large} 
  && 6.07 & 0.66  & 0.08 
  && 4.20 & 1.00  & 0.00  
  && 6.46 & 0.83  & 0.01 
  \\
  && SceneTransformer \cite{ngiam2022scene}
  && 3.99 & 0.49  & 0.11 
  && 3.15 & 0.62  & 0.06  
  && 4.69 & 0.71  & \bfseries 0.04 
  \\
  &&\cellcolor{gray!40}BiFF (Ours) 
  &&\cellcolor{gray!40} 3.74 &\cellcolor{gray!40} \bfseries 0.44  &\cellcolor{gray!40} \bfseries 0.16 
  &\cellcolor{gray!40}&\cellcolor{gray!40} 2.70 &\cellcolor{gray!40} \bfseries 0.53  &\cellcolor{gray!40} 0.06  
  &\cellcolor{gray!40}&\cellcolor{gray!40} 4.42 &\cellcolor{gray!40} \bfseries 0.67  &\cellcolor{gray!40} 0.03
  \\
  &&\cellcolor{gray!40}BiFF-200 (Ours) 
  &&\cellcolor{gray!40} \bfseries 3.64 &\cellcolor{gray!40} 0.46  &\cellcolor{gray!40} \bfseries 0.16 
  &\cellcolor{gray!40}&\cellcolor{gray!40} \bfseries 2.64 &\cellcolor{gray!40} 0.54  &\cellcolor{gray!40} \bfseries 0.07  
  &\cellcolor{gray!40}&\cellcolor{gray!40} \bfseries 4.31 &\cellcolor{gray!40} 0.69  &\cellcolor{gray!40} 0.03 
  \\ 
  \bottomrule
\end{tabular}

\label{tab:leaderboard}
\vspace{-3mm}
\end{table*}

\section{Experiments}
\label{sec: Experiments} 

\subsection{Experimental Setup}

{\bf Dataset and metrics.} 
We train and evaluate our BiFF in the  Waymo Open Motion Dataset (WOMD) \cite{ettinger2021large}, a large-scale dataset with the most diverse interactive cases collected from realistic traffic scenes. We focus on the interactive challenge where the task is to predict 8s future trajectories for two interactive agents given 1s tracking history. The WOMD contains about 487k training samples, 44k validation samples and 44k testing samples. The metrics \cite{ettinger2021large} are calculated with the official tool. Besides, to measure the consistency of the joint trajectory prediction for all $K$ scene modalities, we adopt Cross Collision Rate (CCR) \footnote{Note that the official metric Overlap Rate (OR) in the challenge only calculates collision between prediction and ground truth for the modality with the highest score but not count overlap among joint predicted trajectories and not including other scene modalities } \cite{zhan2019interaction} metric. Concretely, for a modality of a case, the collision is counted if the two agents overlap at any predicted time step. Then the ratio of collision is calculated by $N_{coll} / K$. The final CCR is averaged over all samples. 

{\bf Implementation Details.} Our model is trained for 30 epochs with 4 RTX 3090 GPUs using AdamW optimizer, and the initial learning rate, batch size are set to 0.0001 and 80 respectively. The learning rate is half decayed every 2 epochs from epoch 20. For the scene encoder, we stack $N_E = 6$ layers for the transformer encoder. The map is pruned with marginal heatmaps using half circle and half ellipse to ensure model accurately focusing on the interest area. The number of neighbors $k$ is set to 16 for local self-attention. The hidden feature $d$ is set as 256. For the transformer decoder, we stack $N_H = 1$ layer for HFIF and $N_L = 3$ layers for LFBF. The number of map polylines $L$ is set to 256 for scene context aggregation. The number of static future intentions $S$ is set to 100 as the conditional anchors for HFIF. No data augmentation and model ensemble is used. More details can be referred to appendix.  

\begin{figure}[t]
    \centering
    \includegraphics[width=1.0\columnwidth]{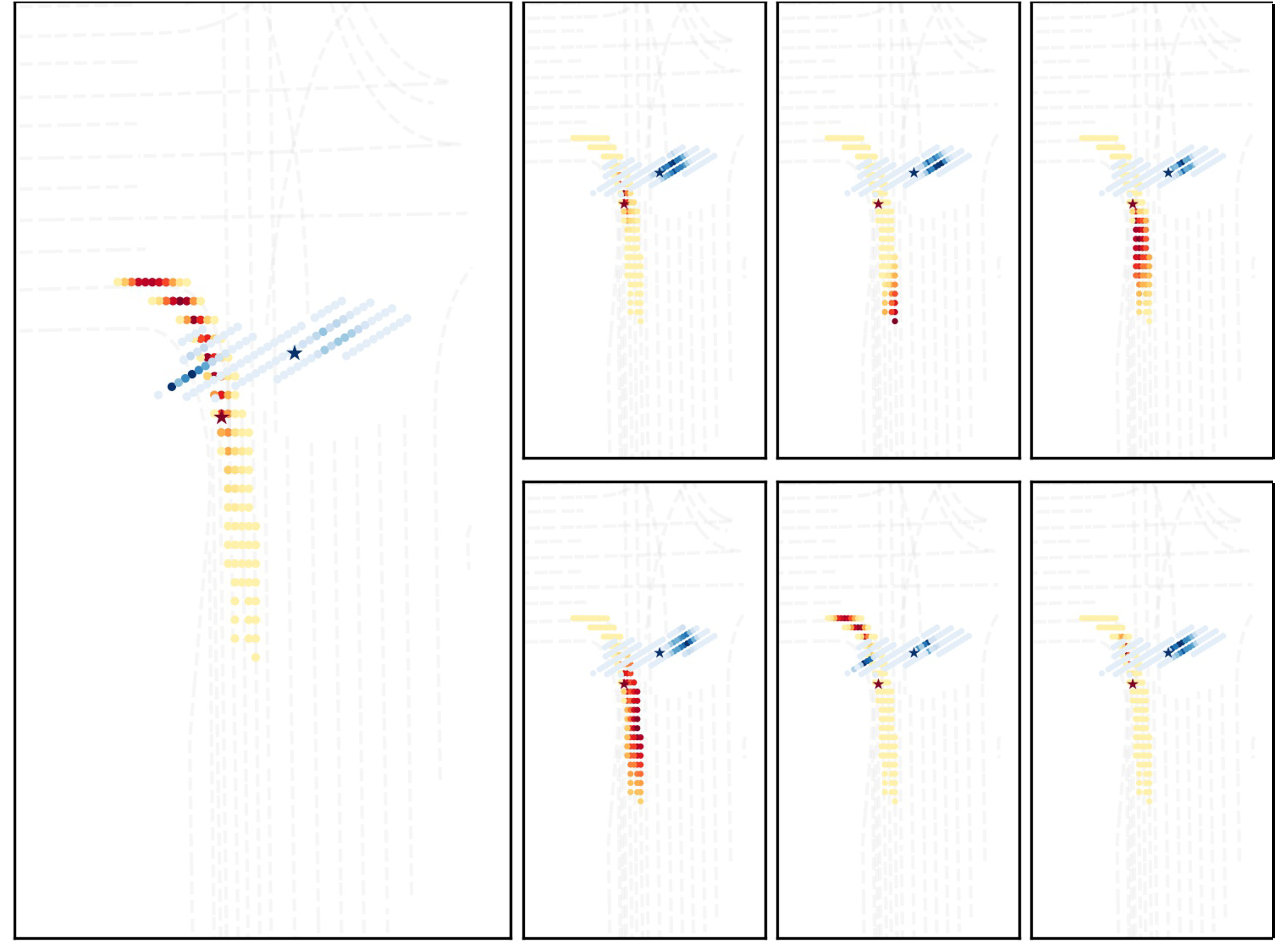}
    \caption{  Illustration of conflict resolution in HFIF model. \textbf{Left}: Marginal heatmaps. \textbf{Right}: Scene-compliant assignment scores from different headers of motion decoder. Brightness in red and blue signifies the score of two interacting agents, with the asterisk denoting the ground truth goal points.}
\label{fig:heatmap}
\end{figure}

\subsection{Comparison with the State-of-the-art}
\begin{figure}[t]
    \centering
     \vspace{-1mm}
    \includegraphics[width=1.0\columnwidth]{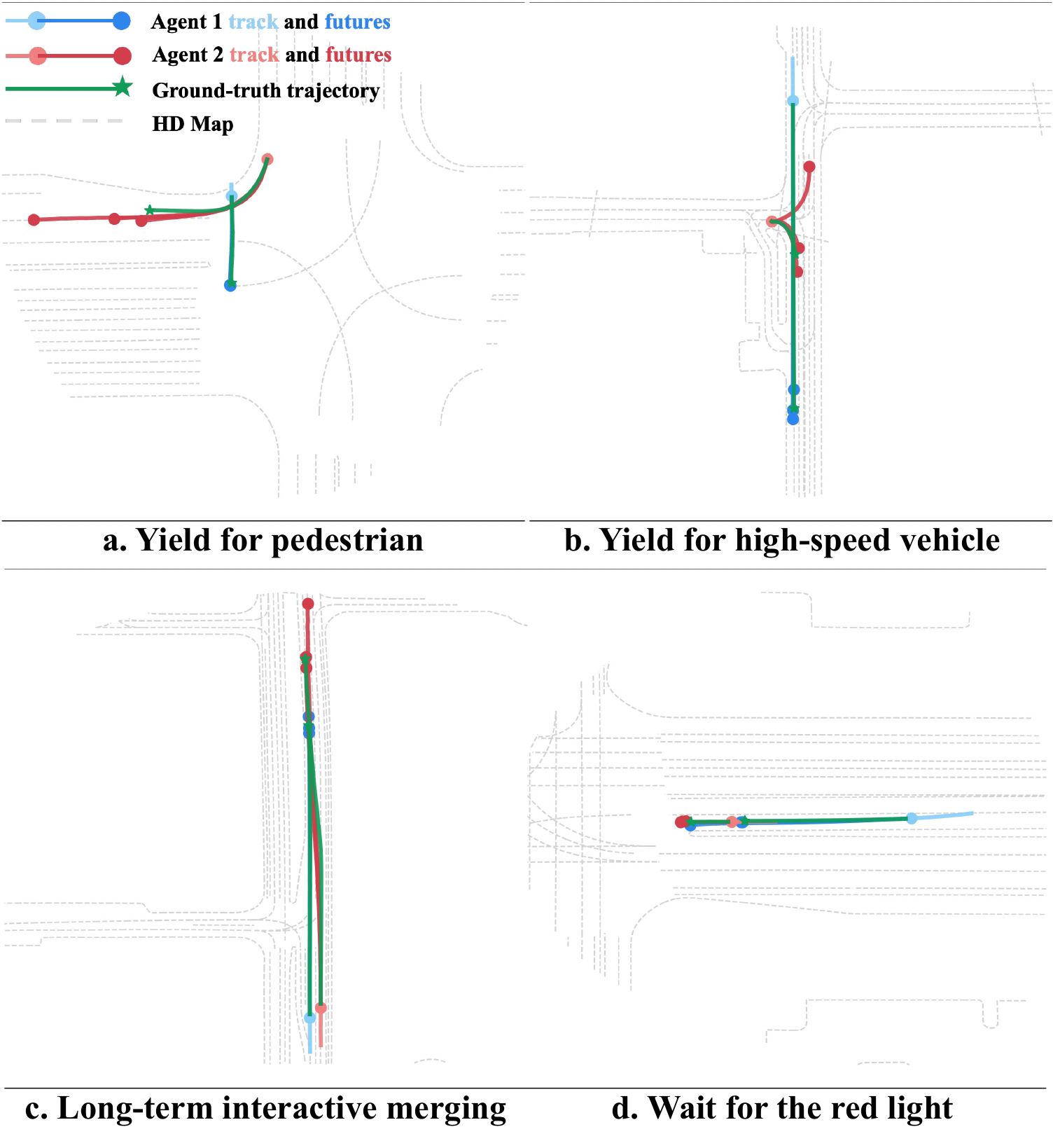}
    
    \caption{  Qualitative results on WOMD validation set (K=3).}
\label{fig:quality_result}

\end{figure}

We compare our BiFF with the SOTA methods on both validation and test interactive WOMD. The quantitative results of proposed BiFF and all baselines are shown in Tab. \ref{tab:leaderboard}, where BiFF-200 denotes BiFF with intention number $S = 200$. For the results in validation set, our BiFF outperforms both Waymo Full Baseline and Scene Transformer in terms of all metrics except for mAP over cyclist. For the results in test set, compared to the approaches without ensemble models, BiFF achieves SOTA in minFDE and MR. Compared to the approaches with ensemble models, BiFF still achieves SOTA in minFDE over vehicle and pedestrian while all other metrics are near SOTA except for mAP. Although BiFF has lower mAP, it improves the minFDE and MR by a large margin over the methods without ensemble models, meaning BiFF predicts more accurate trajectories with efficient scene multi-modality. To illustrate that, we visualize the representative qualitative results in \figref{fig:quality_result}.

\subsection{Ablation Study}
To study the importance of each module in BiFF, we conduct ablation studies on the Waymo interactive validation set over joint metrics of vehicles at 8s. To improve the efficiency of ablation study, we resort to the alternative training scheme with total 15 epochs where the learning rate is initialized as 0.0001 and decayed to half from epoch 10.

{\bf Importance of HFIF.} For simplicity, we name Goal Regression (GR) indicating the transformer decoder in Sec.\ref{sec:HFIF} without HFIF.
For the HFIF-based transformer decoder, illustrated in Tab. \ref{tab:exp_set}, with HFIF module, MR is noticeably improved, which indicates that in interactive scenarios, HFIF plays a crucial role in goal prediction since the future intentions of target agents are likely to be intensely interactive. The future intentions of the interactive counterpart provide useful future information for predicting goals of each target. 

In \figref{fig:heatmap}, an interactive sample is visualized to show the scene-consistent goals generated from multi-modal decoder in HFIF. On the left, the marginal heatmaps (conditional anchors) of two interactive agents are highly multi-modal and conflicting since the peaks are overlapped with each other. Our HFIF is aimed to decompose multi-modal to uni-modal \cite{zhang2022trajectory} for each heatmap while splitting goals across target agents in each scene modality. On the right, we show the assignment scores of both two interactive agents predicted by $K$ different headers. We observe that each header focuses on the distinctive interactive modality and the overlap is successfully eliminated in each scene modality, which verifies that the high-level future intention interaction across target agents is modeled efficiently by our HFIF. 

\begin{table}[ht]
\centering
\setlength{\tabcolsep}{1.0mm}
\ra{1.1}
\footnotesize
\caption{Effects of different components in BiFF. \textbf{GR}: Goal Regression, \textbf{TR}: Trajectory Refine, \textbf{CCR}: Cross Collsion Rate.
}
\vspace{2mm}
\definecolor{forestgreen}{RGB}{47, 159, 87}
\newcommand{\cmark}{\color{forestgreen}\ding{51}}%
\newcommand{\xmark}{\color{red}\ding{55}}%

\begin{tabular}{cccc|cccc}

    \toprule
    \textbf{GR} & \textbf{HFIF} &\textbf{TR} & \textbf{LFBF} &  \textbf{mADE$\downarrow$} &  \textbf{mFDE$\downarrow$} &  \textbf{MR$\downarrow$} &  \textbf{CCR$\downarrow$} \\
    \hline
    \cmark & \xmark & \xmark &\xmark & 2.2292 & 5.0402 &0.6033 &0.1932\\
    \cmark & \cmark & \xmark &\xmark & 2.0167 & 4.5718 &0.5572 &0.1810\\
    \cmark & \cmark & \cmark &\xmark & 1.9283 & 4.3520 &0.5564 &0.1547\\
    \cmark & \cmark & \cmark &\cmark & \bfseries 1.8401 & \bfseries 4.1497 &\bfseries 0.5121 &\bfseries 0.1397\\
    \hline
    
    \end{tabular}
    \label{tab:exp_set}
\end{table}

{\bf Importance of LFBF.} For simplicity, the transformer decoder in Sec.\ref{sec:LFBF} without LFBF is termed as Trajectory Refine (TR).
From Tab. \ref{tab:exp_set}, for the LFBF-based transformer decoder, both components TR and LFBF can improve the performance. First, with TR module, the model can iteratively refine the future trajectory by aggregating the behavior-specific scene context in the future. We find that MR is not improved compared with HFIF, which further verifies the efficiency of HFIF in predicting goals. Moreover, the CCR significantly improves with TR, presumably because the predicted future behavior is more similar to the realistic driving scenario using TR. Thus, TR makes it more reliable for LFBF to perform local future fusion. Second, with LFBF module, the prediction of each target agent can rely on the future location of the interactive counterpart, which is useful since reliable information is strongly associated with future predictions. All metrics are continuously improved, which suggests the importance of LFBF in fusing low-level future behavior across interactive agents.

\begin{figure}[t]
    \centering
    \includegraphics[width=1.0\columnwidth]{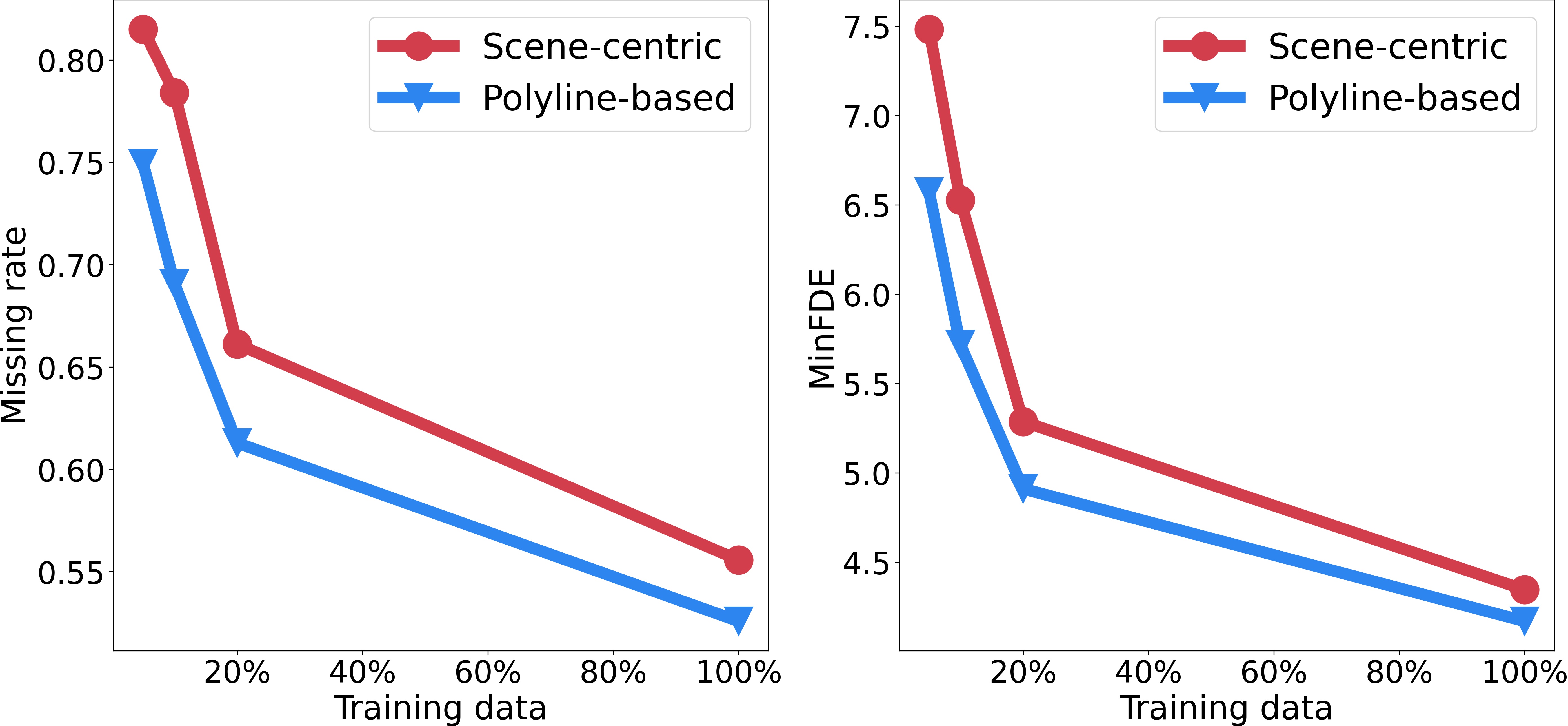}
    
    \caption{ Data efficiency.}
\label{fig:ablation}
\end{figure}

\begin{figure}[t]
\centering
    \includegraphics[width=1.0\linewidth]{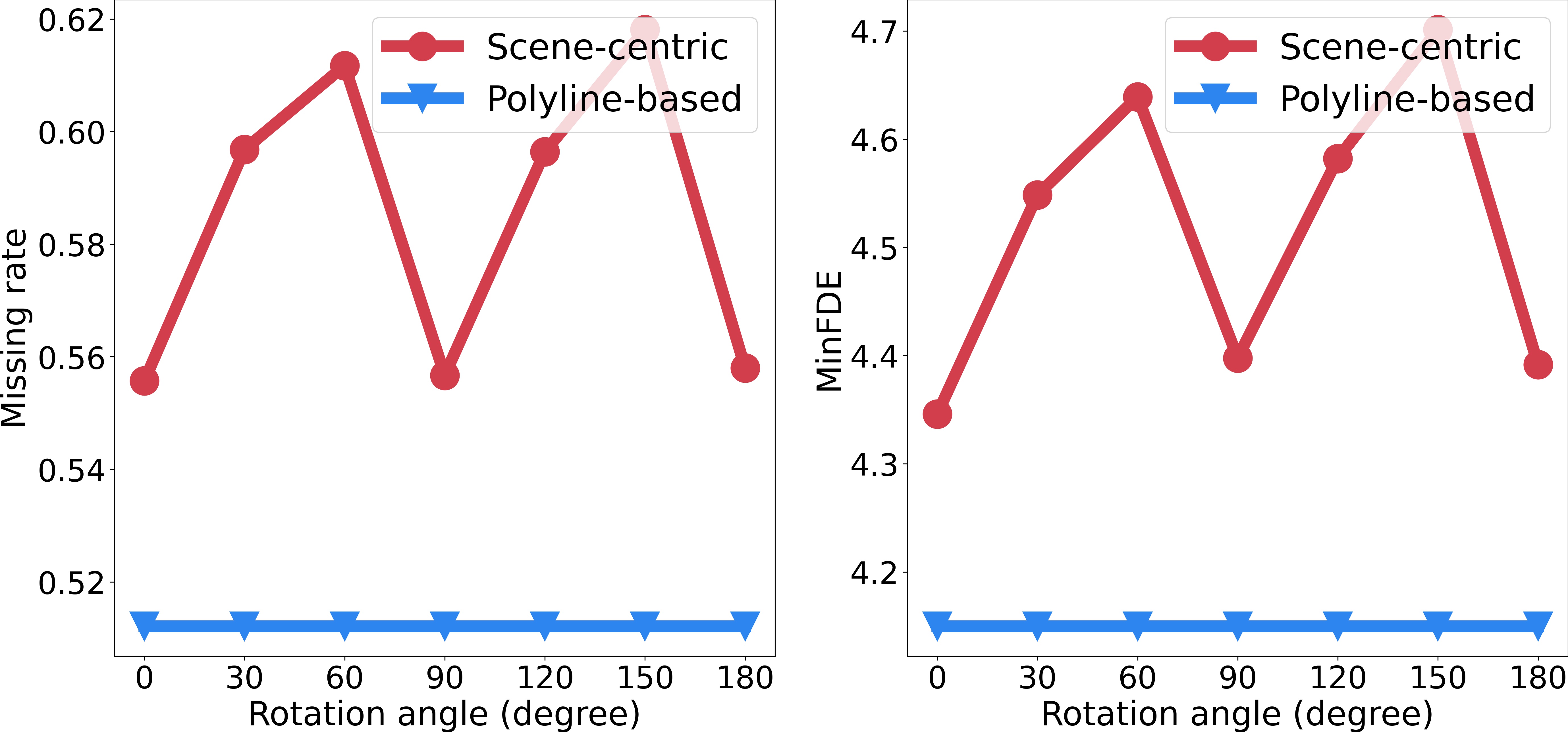}

\caption{
Robustness to rotation angles.
}
\label{fig:appendix_angle}
\end{figure}

{\bf Importance of polyline-based coordinate.} In order to substantiate the efficacy of polyline-based coordinates in enhancing data efficiency and frame robustness, we design a variant of BiFF with scene-centric reference frame, where all scene context is translated to the center location of two target agents at the current time step. To demonstrate data efficiency, we train BiFF and scene-centric BiFF using 5\%, 10\%, 20\% and 100\% frames uniformly sampled from the WOMD training set, and evaluate them on the interactive validation set. From \figref{fig:ablation}, we can find that compared to scene-centric variant, BiFF achieves the same performance with fewer training data. The final prediction accuracy also becomes better, which verifies the improved ability of model generalization with polyline-based coordinate. To show the robustness to coordinate, from Fig. \ref{fig:appendix_angle}, we can observe that our BiFF is robust to the rotation while the performance of the variant severely gets worse with the angle variation. Besides, by normalizing scene-centric BiFF to any polyline-based coordinate, the results become not applicable but the performance of our BiFF is still unchanging, which verifies the robustness to the variant of reference frame by adopting polyline-based coordinate. 

We measure memory efficiency with the Average Polyline Number (APN) for each target agent on the Waymo interactive validation set. By using polyline-based coordinate, the agent-centric strategy is achieved without redundant context, all target agents in each case share the scene context and the APN is 179. However, when encoding local context for each target separately, it causes the redundant context and the APN is 291 meaning requiring much more memory during training and inference.

\begin{table}[ht]
\begin{minipage}[ht]{\columnwidth}
  \caption{Effects of quantity of conditional anchors.}
  \label{tab:anchor num}
  \centering
  \setlength{\tabcolsep}{1.0mm}
\ra{1.1}
\footnotesize
\begin{tabular}{cccc}

    \toprule
    \textbf{Anchor Num} & \textbf{mADE$\downarrow$} &  \textbf{mFDE$\downarrow$} &  \textbf{MR$\downarrow$}\\
    \hline
    50  & 1.9442 &4.4522 &0.5229\\
    100 & 1.8401 &4.1497 &\bfseries 0.5121\\
    200 & \bfseries 1.8284 &\bfseries 4.1058 &0.5434\\
    500 & 1.9405 &4.4488 &0.6009\\
    \hline
    
    \end{tabular}
  \end{minipage}
  \vspace{5mm}
\\[4pt]
\begin{minipage}[!t]{\columnwidth}
  \caption{Effects of quality of conditional anchors.}
  \label{tab:anchor quality}
  \centering
  \setlength{\tabcolsep}{1.0mm}
\ra{1.1}
\footnotesize
\begin{tabular}{cccc}

    \toprule
    \textbf{Anchor Quality (mFDE)} & \textbf{mADE$\downarrow$} &  \textbf{mFDE$\downarrow$} &  \textbf{MR$\downarrow$}\\
    \hline
    4.24                    & 2.1138 &4.9617 &0.6371\\
    3.99                    & 1.9572 &4.4918 &0.5554\\
    3.42 \textbf{w/o} score &\bfseries 1.8401 &\bfseries 4.1497 &\bfseries 0.5121\\
    3.42 \textbf{w} score   & 1.8501 &4.1694 &0.5258\\
    \hline
    
    \end{tabular}
  \end{minipage}
  \vspace{-3mm}
\end{table}

{\bf Analysis of conditional anchors.} We analyze the effects of the quantity and quality of conditional anchors. For the quantity of conditional anchors, by changing the number of intentions in Tab. \ref{tab:anchor num}, we observe that within a certain range, the minFDE is improved with anchor number increasing while the minMR becomes worse.  For the quality of conditional anchors, we define marginal minFDE \footnote{We obtain 6 endpoints by using NMS on the marginal heatmap, and marginal minFDE is obtained by evaluating on the validation set.} to quantify the quality of marginal heatmap (conditional anchors). From Tab. \ref{tab:anchor quality}, all metrics are improved with quality enhancement, which motivates the prediction performance to be further improved by using higher-quality marginal heatmaps. We also find that the performance is slightly dropped by adding the confidence score to each intention, and the marginal score may mislead the model.



\section{Conclusion}
\label{sec:Conclusion}

In this paper, we propose BiFF to model the future interactions, where HFIF fuses high-level future intentions and LFBF fuses low-level future behaviors. With polyline-based coordinates, BiFF becomes data-efficient, robust to global frame and more accurate in prediction. Experimental results demonstrate that BiFF achieves SOTA performance, underscoring the importance of each proposed module.

\paragraph{Acknowledgments.}
We would like to thank Lu Zhang for helpful discussions on the manuscript.

{\small
\bibliographystyle{iccv2023_for_arxiv}
\bibliography{iccv2023_for_arxiv}
}

\clearpage
\appendix
\section*{Appendix}
\section*{A. Implementation Details}

\subsection*{A.1. Predict Marginal Heatmap}
The public source code of M2I \cite{sun2022m2i} is used for predicting marginal heatmap. Specifically, we train three models for vehicle, pedestrian and cyclist separately. To encode scene context, we leverage the context encoder with both vectorized and rasterized representations. Please refer to M2I \cite{sun2022m2i} and its source code for more details.





\subsection*{A.2. Data Preprocessing}

{\bf{Filter Interactive Pairs.}} Apart from the labeled interactive cases in the training set of WOMD, we filter more interactive pairs with the closest spatial distance \cite{sun2022m2i} $d_m$ in the future: 

\begin{equation}
d_m = {\text{min}}_{t_1=1}^T{\text{min}}_{t_2=1}^T{||{\bm y}_1^{t_1}-{\bm y}_2^{t_2}||}_2,
  \label{eq:important}
\end{equation}

where ${\bm y}_1$ and ${\bm y}_2$ are the future trajectories of two agents with $T$ steps. For each training case, we calculate $d_m$ for any pair of agents and iteratively select pair with the smallest $d_m$ over the left agents.  

{\bf{Prune Map with Marginal Heatmap.}} Given the marginal heatmap of each predicted agent, we prune the map limited by the area of top $S$ intentions selected from heatmap using half circle and half ellipse. Concretely, we normalize intentions to the corresponding polyline coordinate of target agent, and calculate the distance between each intention and origin. $d_f$ denotes the maximum distance over the intentions at the front of agent while $d_r$ is the maximum distance over the intentions at the rear of agent. If $d_f > d_r$, the half circle with radius $r = d_f + 30 \text{m}$ is used at the front while the half ellipse with semi-major axis $a=r$ and semi-minor axis $b = d_r + 20 \text{m}$ is used at the rear. If $d_f \le d_r$, the half circle with radius $r = d_r + 30 \text{m}$ is used at the rear while the half ellipse with semi-major axis $a=r$ and semi-minor axis $b = d_f + 20 \text{m}$ is used at the front. For the interactive cases, all road points within the local region of any target agent are reserved for trajectory prediction.

\subsection*{A.3. More Architecture Details}

We train a single model for predicting all types of interactive pairs (any pairwise combination over vehicle, pedestrian and cyclist). Motivated by MTR \cite{shi2022motion}, we use a three-layer MLP with dimension 256 to encode agent polylines,, and use a five-layer MLP with dimension 64 to encode road polylines. Both two types of polylines are further projected to dimension 256 with another linear layer separately. For the multi-modal decoder in HFIF (High-level Future Intentions Fusion), we use 1D convolution for goal regression and a three-layer MLP with dimension 256 for trajectory completion. For the multi-modal decoder in LFBF (Low-level Future Behaviors Fusion), we adopt a three-layer MLP with dimension 512 for trajectory prediction, and the weights are not shared across different layers.

\subsection*{A.4. Inference Latency}

For the default setting of BiFF, the average inference latency is about 56$ms$ for any case from Waymo interactive validation set. We measure the inference latency using a RTX 3090 GPU with standard Pytorch code.






\section*{B. Qualitative Results}

\begin{figure*}[t]
\centering
    \includegraphics[width=1.0\linewidth]{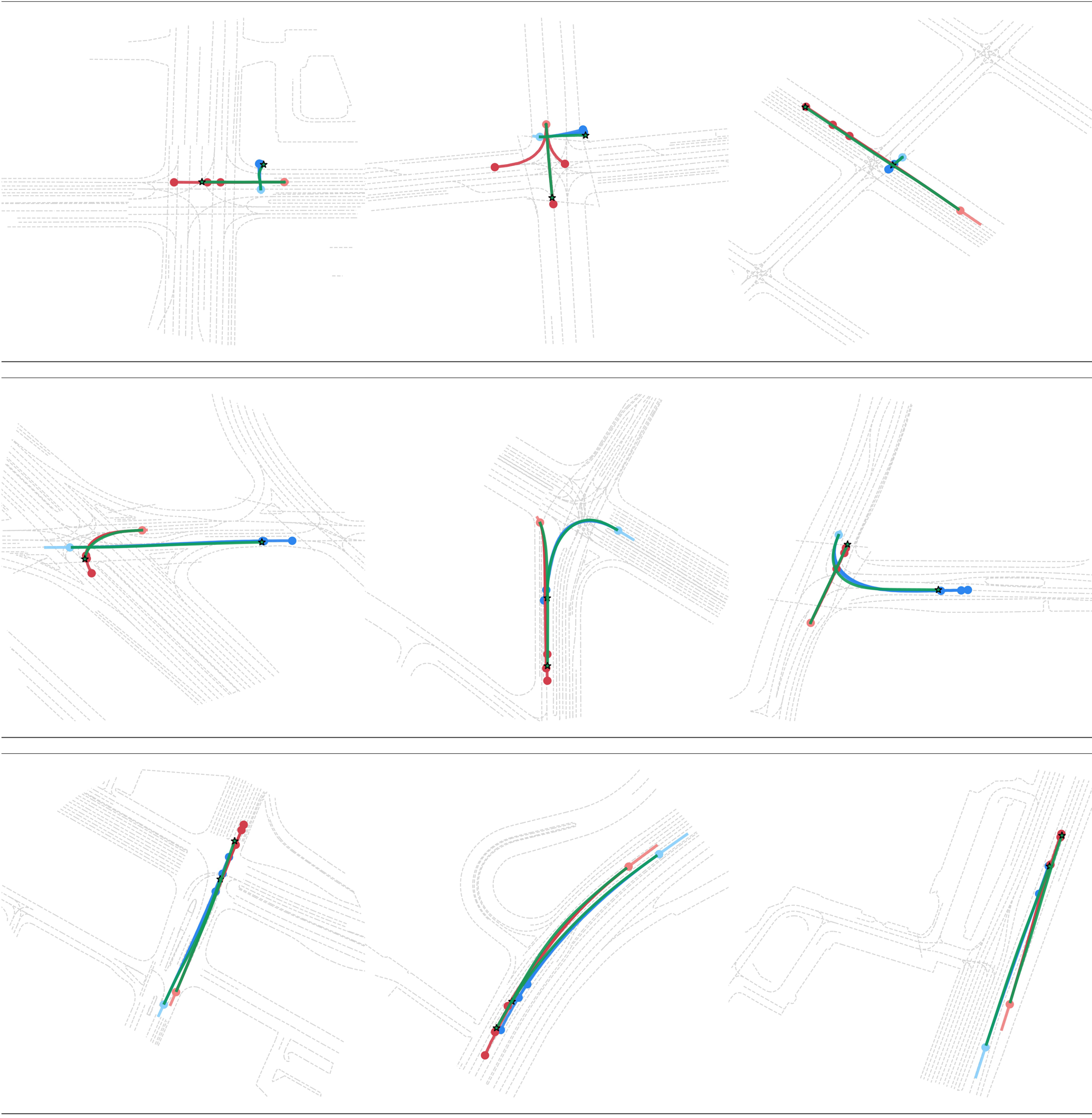}
\caption{
Qualitative results under diverse scenarios on the WOMD interactive validation set. HD map information is shown in light grey. For clarity, we choose K=3 pairs of predicted scene-compliant trajectories shown in red and blue while the corresponding history track is shown in a light color. Ground-truth future trajectories are illustrated in green on the top. 
}
\label{fig:examples}
\vspace{5mm}
\end{figure*}

\begin{figure*}[t]
\centering
    \includegraphics[width=1.0\linewidth]{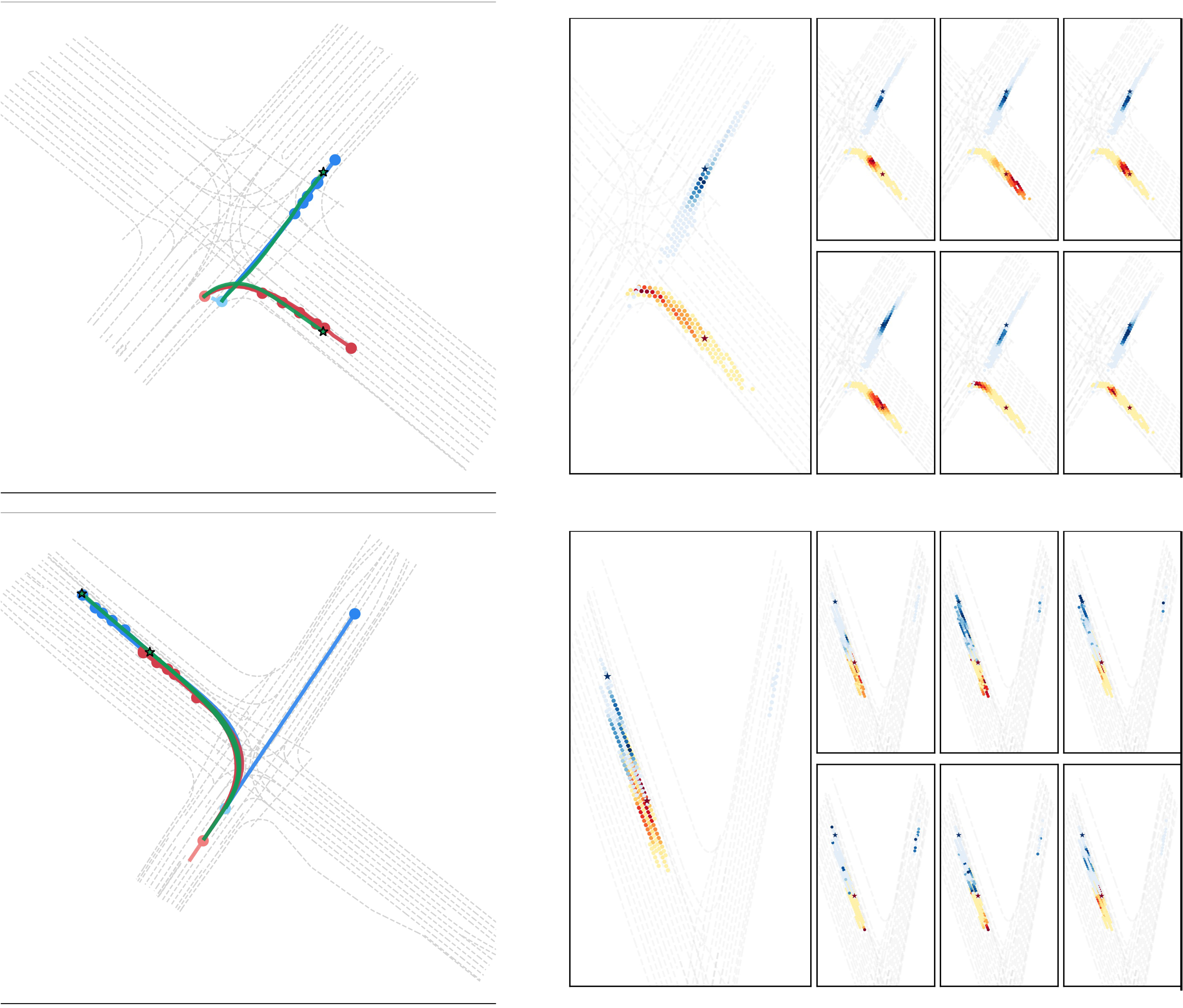}
\vspace{3mm}
\caption{
More qualitative results for conflict resolution. On the left column, we present the map information with predicted trajectories (K=6). On the right, we show the same scenario with heatmap representations. \textbf{Left}: Marginal heatmaps. \textbf{Right}: Scene-compliant assignment scores from different headers of motion decoder. Brightness in red and blue signifies the score of two interacting agents, with the asterisk denoting the ground truth goal points.
}
\label{fig:ablation_heatmap}
\end{figure*}

\begin{figure*}[t]
\centering
    \includegraphics[width=1.0\linewidth]{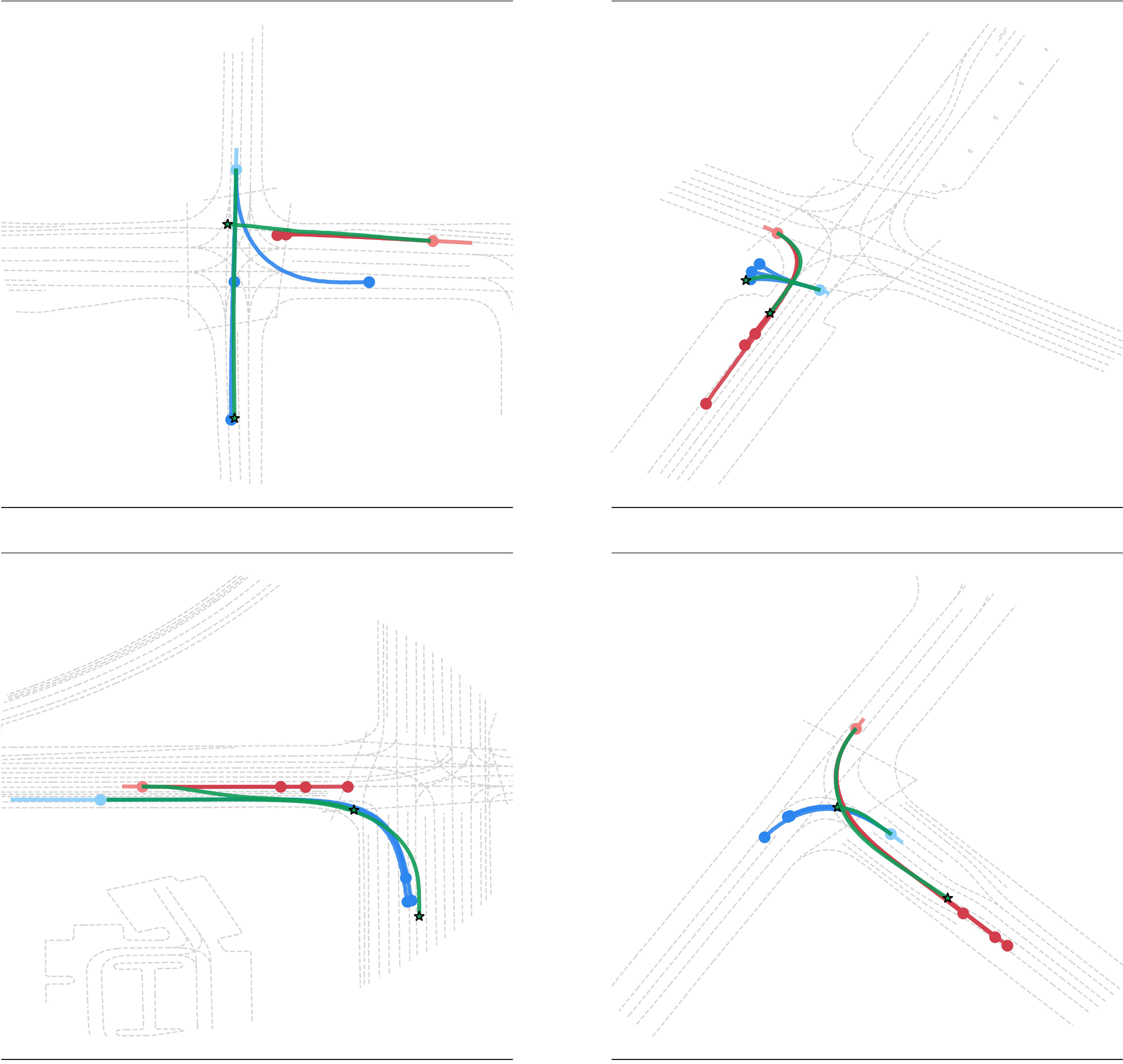}
\vspace{3mm}
\caption{
 Qualitative analysis for failure cases on the WOMD interactive validation set.
}

\label{fig:failure}
\end{figure*}

The visualization results of our proposed method under complex interactive scenarios on the interactive validation set of WOMD are presented in Figure \ref{fig:examples}. The different interactive scenarios are shown in separate rows for clarity. In the first row, we demonstrate the effectiveness of our model in handling various types of agents by showcasing the interactions between vehicles and pedestrians. The middle row presents yielding scenarios among vehicles in complex intersections. Finally, the last row presents interactive merging scenarios where two agents are competing for the right-of-way at high speeds. These results illustrate the ability of our model to accurately predict long-term interactive scenarios in diverse and challenging scenarios. We also present more qualitative results for conflict resolution in Figure \ref{fig:ablation_heatmap}. Finally, we demonstrate the failure cases in Figure \ref{fig:failure}. The failures are mostly related to the misunderstanding of the agent's intention, for example, in the first figure, the blue agent is predicted to turn left in one of the modalities, then the model generates a more conservative behavior for red vehicle to yield the blue one.

\section*{C. Notations}

To illustrate notations in the paper, Tab. \ref{tab:lookup_table} is provided.

\section*{D. Limitation and Future Work}

We have identified several limitations of our proposed BiFF approach and outline potential avenues for future research. First, while BiFF demonstrates promising results, there remains a performance gap between our approach and SOTA in terms of mean average precision (mAP). We hypothesize that this discrepancy may be attributed to the inconsistency between the marginal heatmap and BiFF, as they are trained separately. To address this, one solution is to train a model that predicts the score of each agent separately, supervised with soft labels, and then obtain the joint score by multiplying the scores of all target agents in each modality. Second, limited by computing resources, the current version of BiFF is small without sufficient training data. We anticipate that increasing the amount of training data and decoder layers will continuously enhance the performance of BiFF. Third, the proposed approach can be extended to handle more than two interactive targets with specifically designed techniques like sparse attention to reduce the computation of matrix multiplication.


\section*{E. Broader Impact}

Regarding addressing real-world challenges for autonomous driving beyond the SOTA performance, we list our contributions to advance this area. First, from motivation, the joint trajectory prediction problem addressed by our BiFF is necessary for safe and comfortable driving, since a comprehensive understanding of the future trajectory distribution of all agents is more informative than marginal trajectory prediction. Second, thanks to polyline-based coordinates, our BiFF is memory-efficient without sacrificing accuracy, which is important for practical deployment and real-time inference. Third, apart from prediction task, the simulation and planning in self-driving will benefit from Bi-level Future Fusion by considering prediction as future for agents. Since the predicted trajectories are more scene-consistent with less unrealistic conflicting, BiFF is able to generate naturalistic driving behaviors for simulation and reduce collisions for planning.


\setcounter{table}{4}
\begin{table*}[ht]
\centering
\setlength{\tabcolsep}{1.0mm}
\large
\caption{Lookup table for notations in the paper.
}

\begin{tabular}{cl}

    \toprule
    $A$ & number of predicted interactive agents\\
    $S$ & number of static intentions (conditional anchors)\\
    $D$ & number of hidden feature dimension\\
    $K$ & number of predicted scene modalities\\
    $T$ & number of predicted future steps\\
    $L$ & number of nearest road polylines\\
    ${N}_{E}$ & number of stack layers of transformer encoder with relative positional encoding\\
    ${N}_{L}$ & number of stack layers of LFBF and multi-modal decoder\\
    ${N}_{H}$ & number of stack layers of HFIF\\
    ${P}_{ij}$ & relative positional encoding from polyline $j$ to $i$\\
    ${h}_{i}^{e}$ & features of polyline i\\
    $d$ & the dimension of polyline feature\\
    $\alpha$ & scaled dot-product attention\\
    ${N}_{i}$ & the set of polyline $i$'s neighbors\\
    ${\gamma}^a_{k,s}$ & assignment scores for $a$-th agent in $k$-the modality \\
    ${L}_{G}$ & Goal regression\\
    ${L}_{T}$ & Trajectory regression\\
    $\mathbf{y}^{1:T}$ & Predicted future trajectory \\
    ${N}_{coll}$ & number of collision when calculating Cross Collision Rate\\

    \end{tabular}
    \label{tab:lookup_table}

\end{table*}

\end{document}